\pdfoutput=1

\documentclass[11pt,a4paper]{article}
\usepackage{times,latexsym}
\usepackage{url}
\usepackage[T1]{fontenc}

\usepackage[acceptedWithA]{tacl2021v1}

\usepackage{xspace,mfirstuc,tabulary}

\newif\iftaclinstructions
\taclinstructionsfalse 
\iftaclinstructions
\renewcommand{\confidential}{}
\renewcommand{\anonsubtext}{(No author info supplied here, for consistency with
TACL-submission anonymization requirements)}
\newcommand{\instr}
\fi

\iftaclpubformat 

\else

\fi

\usepackage{times}
\usepackage{latexsym}

\usepackage[T1]{fontenc}

\usepackage[utf8]{inputenc}

\usepackage{microtype}

\usepackage{inconsolata}

\usepackage[utf8]{inputenc} 
\usepackage[T1]{fontenc}    
\usepackage{hyperref}       
\usepackage{url}            
\usepackage{booktabs}       
\usepackage{amsfonts}       
\usepackage{nicefrac}       
\usepackage{microtype}      
\usepackage{lipsum}		
\usepackage{graphicx}
\usepackage{natbib}
\usepackage{doi}

\usepackage{amsmath}
\usepackage{amssymb}
\usepackage{mathtools}
\usepackage{amsthm}

\usepackage{multirow}
\usepackage{graphics}
\usepackage{multicol}
\usepackage{lipsum}
\usepackage{mwe}
\usepackage{url}
\usepackage{xcolor}
\usepackage[ruled,vlined]{algorithm2e}
\usepackage{bbm}
\usepackage{upgreek}
\usepackage{subcaption}

\SetAlFnt{\small}
\usepackage{wrapfig}
\usepackage{colortbl}

\usepackage{amsmath,amsfonts,bm}

\def\eqref#1{equation~\ref{#1}}

\def\1{\bm{1}}

\DeclareMathAlphabet{\mathsfit}{\encodingdefault}{\sfdefault}{m}{sl}
\SetMathAlphabet{\mathsfit}{bold}{\encodingdefault}{\sfdefault}{bx}{n}

\DeclareMathOperator*{\argmax}{arg\,max}

\newcommand{\authorA}[1]{{\color{blue} [{\bf authorA}: #1]}}
\newcommand{\authorB}[1]{{\color{red} [{\bf authorB}: #1]}}

\newcommand{\method}{\textsc{FRAME}}
\newcommand{\methodsp}{\textsc{FRAME}\xspace}
\newcommand{\eg}{\textit{e.g., }}
\newcommand{\ie}{\textit{i.e., }}
\newcommand{\versus}{\textit{vs. }}

\colorlet{lred}{red!15}
\colorlet{lblue}{blue!15}
\colorlet{lgreen}{green!20}

\DeclareUnicodeCharacter{FFFD}{ }

\title{\method: Evaluating Rationale-Label \\ Consistency Metrics for Free-Text Rationales}

\date{} 					

\author{
    \textbf{Aaron Chan}$^{\clubsuit}$\thanks{~~Work done while AC was a Meta AI student researcher.} \hspace{4mm} \textbf{Shaoliang Nie}$^{\diamondsuit}$ \hspace{3mm} \textbf{Liang Tan}$^{\diamondsuit}$ \hspace{3mm} \textbf{Xiaochang Peng}$^{\diamondsuit}$ \\ \textbf{Hamed Firooz}$^{\diamondsuit}$ \hspace{3mm} \textbf{Maziar Sanjabi}$^{\diamondsuit}$ \hspace{3mm} \textbf{Xiang Ren}$^{\clubsuit}$ \vspace{1mm} \\ 
    $^{\clubsuit}$University of Southern California \hspace{1mm} $^{\diamondsuit}$Meta AI \vspace{1mm} \\
    \small{\texttt{\{chanaaro, xiangren\}@usc.edu}} \\ 
    \small{\texttt{\{snie, liangtan, xiaochang, mhfirooz, maziars\}@fb.com}}
}

\begin{document}
\maketitle

\begin{abstract}
\renewcommand{\authorA}[1]{}
\renewcommand{\authorB}[1]{}


Following how humans communicate, free-text rationales aim to use natural language to explain neural language model (LM) behavior.
However, free-text rationales' unconstrained nature makes them prone to hallucination, so it is important to have metrics for free-text rationale quality.
Existing free-text rationale metrics measure how consistent the rationale is with the LM's predicted label, but there is no protocol for assessing such metrics' reliability.
Thus, we propose \method, a framework for evaluating rationale-label consistency (RLC) metrics for free-text rationales.
\methodsp is based on three axioms: (1) good metrics should yield highest scores for reference rationales, which maximize RLC by construction; (2) good metrics should be appropriately sensitive to semantic perturbation of rationales; and (3) good metrics should be robust to variation in the LM's task performance.
Across three text classification datasets, we show that existing RLC metrics cannot satisfy all three \methodsp axioms, since they are implemented via model pretraining which muddles the metric's signal.
Then, we introduce a non-pretraining RLC metric that greatly outperforms baselines on (1) and (3), while performing competitively on (2).
Finally, we discuss the limitations of using RLC to evaluate free-text rationales.

\end{abstract}


\renewcommand{\authorA}[1]{}
\renewcommand{\authorB}[1]{}

\section{Introduction} 
\label{sec:intro}


\begin{figure}[t!]
\centering
\includegraphics[width=0.7\linewidth]{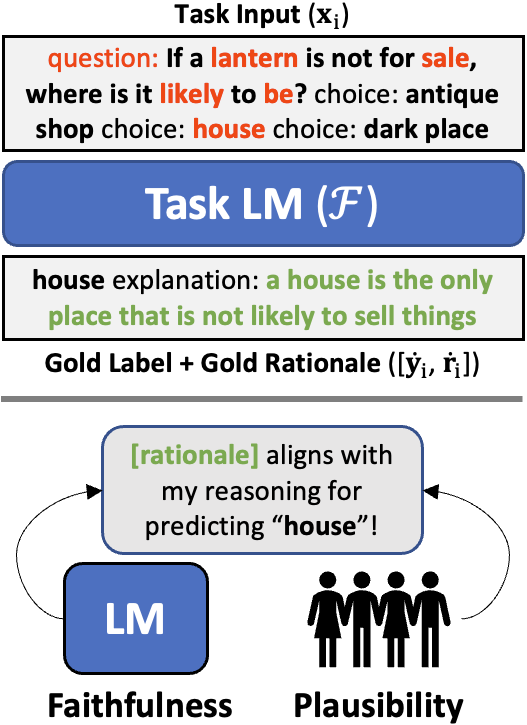}
\caption{\small \textbf{Free-Text Rationales.} Whereas extractive rationales (red) highlight input tokens, \textit{free-text rationales} (green) use natural language to explain a task LM's ($\mathcal{F}$) label prediction. Given $\mathbf{x}_i$ as input, a self-rationalizing $\mathcal{F}$ is trained to jointly generate the gold label ($\mathbf{\dot{y}}_i$ = \texttt{house}) and gold rationale ($\mathbf{\dot{r}}_i$ = \texttt{a house ... sell things}) as a single output sequence. Good free-text rationales should be both \textit{faithful} and \textit{plausible}.}
\label{fig:f_model}
\vspace{-0.4cm}
\end{figure}

\begin{figure}[t!]
\centering
\includegraphics[width=0.6\linewidth]{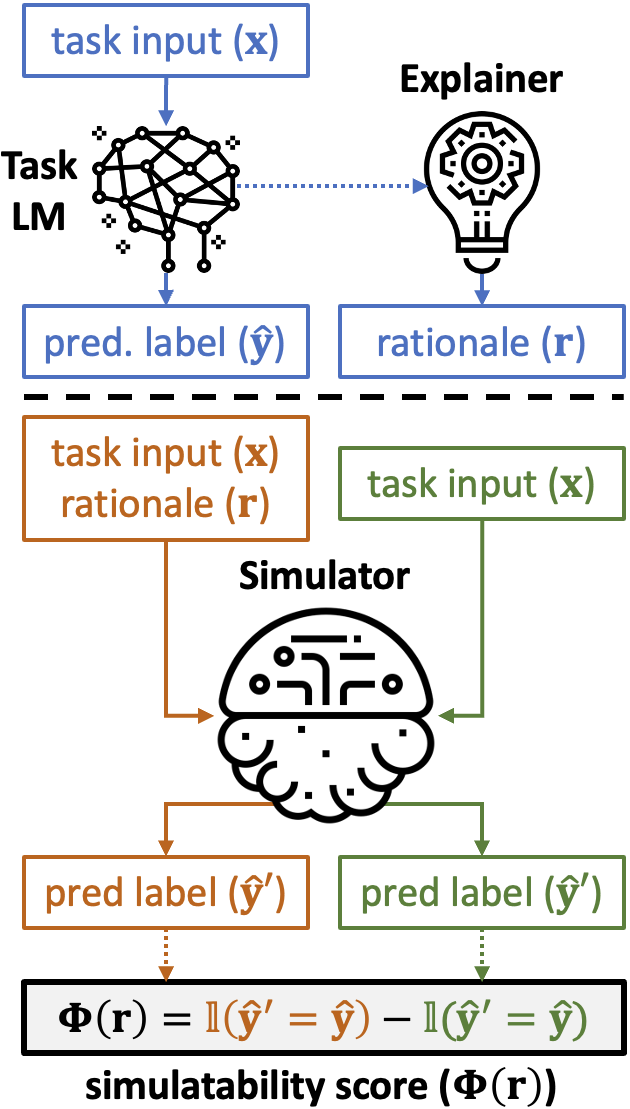}
\caption{\small \textbf{Evaluating Rationales via RLC/Sim.} Given task input $\mathbf{r}_i$ and the task LM's predicted label $\mathbf{\hat{y}}_i$, we can measure rationale ($\mathbf{r}_i$) quality based on \textit{rationale-label consistency} (RLC). RLC is implemented using \textit{simulatability} (sim), \ie how accurately a simulator (\ie LM or human observer) predicts $\mathbf{\hat{y}}_i$ using both task $\mathbf{x}_i$ and $\mathbf{r}_i$ as input, compared to using only $\mathbf{x}_i$.
}
\vspace{-0.4cm}
\label{fig:sim}
\end{figure}


Neural language models (LMs) perform well on many NLP tasks \citep{devlin2018bert, liu2019roberta}, but their complex behavior can be hard to explain \citep{rudin2019stop, lipton2018mythos, bender2021dangers}.
Unlike extractive rationales \citep{denil2014extraction, sundararajan2017axiomatic, chan2022unirex} which are limited to input token scoring, \textit{free-text rationales} are designed to explain LMs' decisions in a human-like manner via natural language and can describe things beyond the task input (Fig. \ref{fig:f_model}) \citep{narang2020wt5, lakhotia2020fid, kumar2020nile, rajani2019explain, camburu2018snli, park2018multimodal}.
Even so, free-text rationales' unconstrained nature makes them prone to hallucination \citep{ji2022survey, kumar2020nile, wu2018faithful}, so it is important to have metrics for free-text rationale quality.

Given an LM's (\ie task LM's) predicted class label and a rationale for this prediction, existing free-text rationale metrics aim to measure how consistent the rationale is with the predicted label \cite{hase2020leakage, wiegreffe2021measuring}.
In practice, this notion of \textit{rationale-label consistency} (RLC) is implemented using \textit{simulatability} (sim) (Fig. \ref{fig:sim}): How much does the rationale improve a \textit{simulator}'s (\ie observer's) accuracy in predicting the task LM's predicted label \cite{doshi2017towards, hase2020evaluating}?
It follows that sim (\ie RLC) at least partially reflects how aligned the rationale is with the simulator's reasoning process for reaching the predicted label.
Therefore, as the simulator can be an LM or human \citep{hase2020leakage}, RLC can be seen as a prerequisite for key rationale desiderata like \textit{faithfulness} (alignment with LM's reasoning) \citep{jacovi2020towards} and \textit{plausibility} (alignment with humans' reasoning) \citep{deyoung2019eraser}.
Though RLC may not fully capture these desiderata, it may still provide useful signal about them.
However, there is no standard protocol for evaluating RLC metrics' reliability, so their effectiveness remains unclear.


In light of this, we propose \textbf{F}ree-Text \textbf{R}ationale-L\textbf{A}bel Consistency \textbf{M}eta-\textbf{E}valuation (\textbf{\method}), a framework for evaluating RLC metrics for free-text rationales.
\methodsp is based on three axioms: (1) good metrics should yield highest scores for \textit{reference rationales}, which maximize RLC by construction; (2) good metrics should be appropriately sensitive to semantic perturbation of rationales; and (3) good metrics should be robust to variation in the task LM's task performance.
For (1), we simply define the reference rationale as the task LM's predicted label.
This provides a powerful invariance for analyzing the quality of free-text rationales, whose unconstrained nature typically makes such analysis difficult.
For (2), we test if the metric responds appropriately to equivalent (should not change meaning) and contrastive (should change meaning) perturbations of reference rationales.
For (3), we compute RLC as a function of the task LM's task performance, by varying factors like the task LM's number of train instances, number of noisy train instances, and capacity.
We also separately compute RLC on correctly and incorrectly predicted test instances, to check if it is stable across these two subpopulations.

On three text classification datasets---e-SNLI, CoS-E v1.0, and CoS-E v1.11---we show that existing RLC metrics cannot satisfy all three \methodsp axioms.
Since they involve pretraining simulators on external corpora, existing RLC metrics struggle to isolate the rationale's contribution to simulator accuracy from the pretraining knowledge's, hence muddling the metric's signal.
Thus, we introduce a non-pretraining RLC metric, \textsc{NP-$\mathcal{GH}$-Pred}, to address this issue.
For LM simulators, \textsc{NP-$\mathcal{GH}$-Pred} improves performance on (1) and (3) by an average of 41.7\% and 42.9\%, respectively, while performing competitively on (2) (Sec. \ref{sec:exp:faith}).
For human simulators, we conduct a user study of (1), with \textsc{NP-$\mathcal{GH}$-Pred} improving performance on (1) by 47.7\% (Sec. \ref{sec:exp:plaus}).
Based on these results, we discuss the inherent limitations of using RLC to measure free-text rationale quality (Sec. \ref{sec:conclusion}).
\section{Free-Text Rationales} 
\label{sec:background}


Let $\mathcal{F}_{\text{task}}$ denote a task LM for $m$-class text classification. 
For task instance $i$, let $\mathbf{x}_i = [x_{i}^{t}]_{t=1}^{n_x}$ be the $n_x$-token input sequence (\eg a sentence), $\dot{y}_i$ be the gold label, and $\hat{y}_i$ be $\mathcal{F}_{\text{task}}$'s predicted label.
Let $\mathbf{\dot{y}}_i = [\dot{y}_{i}^{t}]_{t=1}^{n_{\dot{y}}}$ and $\mathbf{\hat{y}}_i = [\hat{y}_{i}^{t}]_{t=1}^{n_y}$ be token sequence representations of $\dot{y}_i$ and $\hat{y}_i$, respectively.
A \textit{free-text rationale} $\mathbf{r}_i = [r_{i}^{t}]_{t=1}^{n_r}$ is an $n_r$-token sequence that uses natural language to explain the reasoning process behind $\mathcal{F}_{\text{task}}$ predicting $\hat{y}_i$ (Fig. \ref{fig:f_model}-\ref{fig:h_model}).

\subsection{Free-Text Rationale Generation}
\label{sec:background:gen}
Unlike extractive rationales, which are constrained to scoring tokens in $\mathbf{x}_i$, free-text rationales can have arbitrary content, style, and length (\ie $n_r$) \citep{narang2020wt5, camburu2018snli}.
Free-text rationales can be generated via any means, but are typically generated by an explainer LM $\mathcal{F}_{\text{expl}}$.
In general, we use $\mathbf{r}_i$ to denote a free-text rationale with no assumptions about its generation process, but we specifically denote $\mathcal{F}_{\text{expl}}$'s predicted rationale as $\mathbf{\hat{r}}_i$.
While $\mathcal{F}_{\text{task}}$ and $\mathcal{F}_{\text{expl}}$ can be separate \citep{kumar2020nile, rajani2019explain, camburu2018snli}, \textit{self-rationalizing LMs} combine $\mathcal{F}_{\text{task}}$ and $\mathcal{F}_{\text{expl}}$ into a single LM $\mathcal{F}$, which jointly generates the task and rationale outputs \citep{narang2020wt5, lakhotia2020fid, do2020snli, liu2018towards} (Fig. \ref{fig:f_model}).
In particular, text-to-text self-rationalizing LMs (\eg T5 \cite{raffel2019exploring}) output an ($n_y+n_r$)-token sequence $[\mathbf{\hat{y}}_i, \mathbf{\hat{r}}_i]$.
Following recent works \citep{hase2020leakage, wiegreffe2021measuring}, we focus on T5-based self-rationalizing LMs as $\mathcal{F}$.


\subsection{Free-Text Rationale Evaluation}
\label{sec:background:eval}



Given $\mathbf{x}_i$ and $\mathcal{F}$, existing free-text rationale metrics evaluate $\mathbf{r}_i$ by measuring the consistency between $\mathbf{r}_i$ and $\mathbf{\hat{y}}_i$, denoted as \textit{rationale-label consistency} (RLC) or $\rho(\mathbf{r}_i, \mathbf{\hat{y}}_i, \mathbf{x}_i; \mathcal{F})$.
$\rho$ is computed via \textit{simulatability} (sim), which measures how predictive $\mathbf{r}_i$ is of $\mathbf{\hat{y}}_i$ \citep{doshi2017towards, hase2020evaluating}.
That is, how accurately does a \textit{simulator} (\ie observer) predict $\mathbf{\hat{y}}_i$ using both $\mathbf{x}_i$ and $\mathbf{r}_i$ as input, compared to using only $\mathbf{x}_i$?
It follows that sim (\ie RLC) at least partially reflects how aligned $\mathbf{r}_i$ is with the simulator's reasoning process for reaching $\mathbf{\hat{y}}_i$.
Let $\Phi$ be a RLC metric and $\mathcal{S}$ be a simulator.
Then, $\Phi(\mathbf{r}_i, \mathbf{\hat{y}}_i, \mathbf{x}_i; \mathcal{F}, \mathcal{S}) = \mathbbm{1}_{\mathcal{S}}(\mathbf{\hat{y}}_i | \mathbf{x}_i, \mathbf{r}_i) - \mathbbm{1}_{\mathcal{S}}(\mathbf{\hat{y}}_i | \mathbf{x}_i)$, where $0 \leq \mathbbm{1}_{c}(b \hspace{0.5mm} | \hspace{0.5mm} a) \leq 100$ is $c$'s accuracy in predicting $b$ given $a$.
$\mathbbm{1}_{\mathcal{S}}(\mathbf{\hat{y}}_i | \mathbf{x}_i)$ and $\mathbbm{1}_{\mathcal{S}}(\mathbf{\hat{y}}_i | \mathbf{x}_i, \mathbf{r}_i)$ are the \textit{control} and \textit{treatment} terms, respectively, with higher $\Phi$ indicating higher $\rho$ (\ie $\mathbf{r}_i$ quality).
The rationale desideratum (\eg faithfulness, plausibility) addressed by $\Phi$ is determined by the choice of $\mathcal{S}$ (\eg LM, human).
We discuss this below.





\paragraph{LM Simulators}
For faithfulness, we can set $\mathcal{S}$ as another LM \cite{hase2020leakage, wiegreffe2021measuring}, such that $\Phi$ reflects $\mathbf{r}_i$'s alignment with the LM's reasoning process for reaching $\mathbf{\hat{y}}_i$.
In doing so, prior works implicitly make two strong assumptions.
First, they assume $\rho$ is sufficiently informative about $\mathcal{F}$'s reasoning process w.r.t. $\mathbf{r}_i$, $\mathbf{\hat{y}}_i$, and $\mathbf{x}_i$.
Although a rationale that maximizes $\rho$ may not necessarily explain $\mathcal{F}$'s entire reasoning process, such a rationale may still provide useful signal about $\mathcal{F}$'s behavior.
Second, despite them being different LMs, they assume $\mathcal{S}$'s reasoning process is sufficiently similar to $\mathcal{F}$'s w.r.t. the given task, such that $\mathcal{S}$'s behavior reflects $\mathcal{F}$'s.
This is needed because using $\mathcal{F}$ as the simulator would trivially result in perfect $\mathbf{\hat{y}}_i$ prediction accuracy for the control term, rendering $\Phi$ meaningless.

\begin{figure}[t!]
\centering
\includegraphics[width=0.8\linewidth]{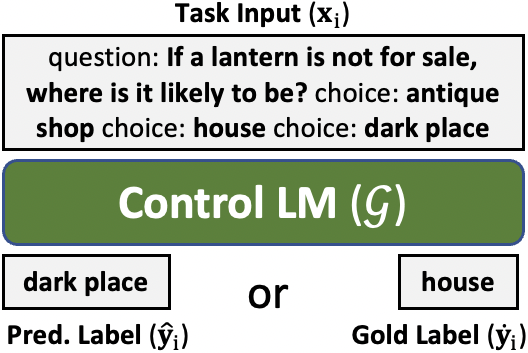}
\caption{\small \textbf{Control LM.} Given task input $\mathbf{x}_i$, the control LM ($\mathcal{G}$) is an LM simulator trained to generate either $\mathcal{F}$'s predicted label $\mathbf{\hat{y}}_i$ or the gold label $\mathbf{\dot{y}}_i$. 
}
\label{fig:g_model}
\end{figure}

\begin{figure}[t!]
\centering
\includegraphics[width=\linewidth]{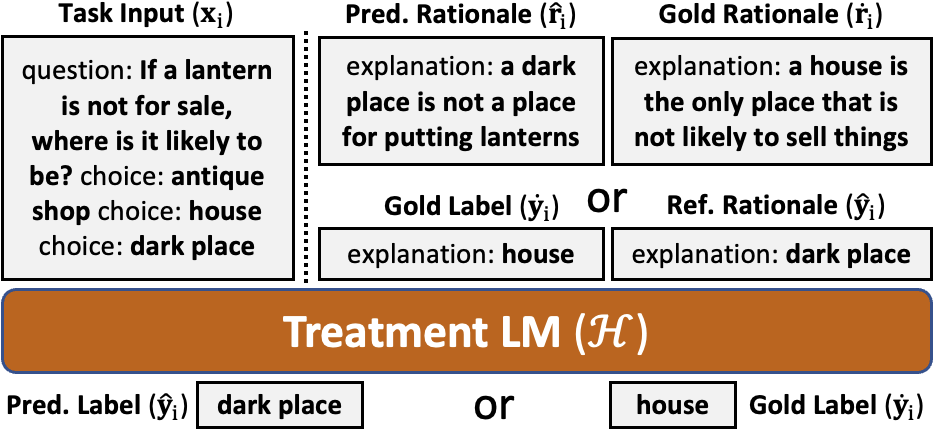}
\caption{\small \textbf{Treatment LM.} Given task input $\mathbf{x}_i$ and some rationale (one of $\mathbf{r}_i$, $\mathbf{\dot{y}}_i$, $\mathbf{\dot{r}}_i$, or $\mathbf{\hat{y}}_i$), the treatment LM ($\mathcal{H}$) is an LM simulator trained to generate either $\mathcal{F}$'s predicted label $\mathbf{\hat{y}}_i$ or the gold label $\mathbf{\dot{y}}_i$.}
\label{fig:h_model}
\vspace{-0.4cm}
\end{figure}

In practice, it can be difficult for $\mathcal{S}$ to jointly approximate the $P(\hspace{0.5mm} \cdot \hspace{0.5mm} |\mathbf{x}_i)$ and $P( \hspace{0.5mm} \cdot \hspace{0.5mm} | \mathbf{x}_i, \mathbf{r}_i)$ distributions.
To address this issue, \citet{wiegreffe2021measuring} decomposes $\mathcal{S}$ into two LMs: $\mathcal{G}$ for $P( \hspace{0.5mm} \cdot \hspace{0.5mm} | \mathbf{x}_i)$ and $\mathcal{H}$ for $P(\hspace{0.5mm} \cdot \hspace{0.5mm} |\mathbf{x}_i, \mathbf{r}_i)$.
Meanwhile, \citet{hase2020leakage} trains $\mathcal{S}$ to approximate both $P(\hspace{0.5mm} \cdot \hspace{0.5mm} |\mathbf{x}_i)$ and $P( \hspace{0.5mm} \cdot \hspace{0.5mm} | \mathbf{x}_i, \mathbf{r}_i)$ by randomly dropping out either $\mathbf{x}_i$ or $\mathbf{r}_i$ from the input during training.
In this paper, we follow the first approach, since it enables us to use the same standard training process for all LMs.
We call $\mathcal{G}$ (Fig. \ref{fig:g_model}) and $\mathcal{H}$ (Fig. \ref{fig:h_model}) the \textit{control LM} and \textit{treatment LM}, respectively.
Then, $\Phi$ is computed as: $\Phi(\mathbf{r}_i, \mathbf{\hat{y}}_i, \mathbf{x}_i; \mathcal{F}, \mathcal{G}, \mathcal{H}) = \mathbbm{1}_{\mathcal{H}}(\mathbf{\hat{y}}_i | \mathbf{x}_i, \mathbf{r}_i) - \mathbbm{1}_{\mathcal{G}}(\mathbf{\hat{y}}_i | \mathbf{x}_i)$.
Unless $\Phi$ is specifically not defined w.r.t. $\mathbf{\hat{y}}_i$, $\mathbf{x}_i$, $\mathcal{F}$, $\mathcal{G}$, and $\mathcal{H}$, we abbreviate $\Phi(\mathbf{r}_i, \mathbf{\hat{y}}_i, \mathbf{x}_i; \mathcal{F}, \mathcal{G}, \mathcal{H})$ as $\Phi(\mathbf{r}_i)$. 

\paragraph{Human Simulators}
For plausibility, $\mathcal{S}$ is a human, such that $\Phi$ reflects $\mathbf{r}_i$'s alignment with humans' reasoning process for reaching $\mathbf{\hat{y}}_i$ \cite{doshi2017towards, hase2020evaluating, chan2022unirex}.
Note that sim was originally defined in the context of plausibility \cite{doshi2017towards}, before being adapted for faithfulness \cite{hase2020leakage}. 
Like in faithfulness, this implicitly assumes all humans share a sufficiently similar reasoning process w.r.t. the given task.
We use the same $\mathcal{S}$ for both $P(\hspace{0.5mm} \cdot \hspace{0.5mm} |\mathbf{x}_i, \mathbf{r}_i)$ and $P( \hspace{0.5mm} \cdot \hspace{0.5mm} | \mathbf{x}_i)$, since a human $\mathcal{S}$ generally does not require fine-tuning.

\paragraph{Issues with RLC Metrics}
Intuitively, RLC metrics cannot fully capture faithfulness or plausibility, as RLC provides only a narrow view of LMs' or humans' reasoning processes.
Also, we see that using RLC for this narrow faithfulness or plausibility evaluation still requires relatively strong assumptions.
Since RLC is currently the only available tool for evaluating rationale quality, one may be inclined to use RLC metrics by default.
However, there exists no protocol for evaluating RLC metrics themselves, so it is unclear how reliable they are.

\section{\method} 
\label{sec:method}


In Sec. \ref{sec:background:eval}, we defined $\Phi$ as a metric for evaluating rationale quality via RLC (\ie $\rho$).
Ideally, $\Phi$ should capture only $\rho$, without entangling other orthogonal information (\eg $\mathcal{F}$'s task performance).
While a number of $\Phi$ metrics have been proposed \cite{hase2020leakage, wiegreffe2021measuring}, there is no standard protocol for evaluating $\Phi$, so it is unclear if such metrics are behaving as desired.
Thus, we propose \method, a general framework for evaluating $\Phi$.
\methodsp assesses $\Phi$ according to three axioms: (1) good metrics should yield highest scores for reference rationales, which maximize RLC by construction; (2) good metrics should be appropriately sensitive to semantic perturbation of rationales; and (3) good metrics should be robust to variation in the task LM's task performance.
\methodsp can be used to evaluate $\Phi$ w.r.t. both LM and human simulators.
Still, in this work, we mainly focus on LM simulators, since they are fully automated.
Sec. \ref{sec:method:axioms} introduces each axiom and its corresponding meta-metrics for $\Phi$, Sec. \ref{sec:method:implementation} provides implementation details for each axiom's meta-metrics, and Sec. \ref{sec:method:sim_metrics} describes all RLC metrics considered in the paper.

\subsection{\methodsp Axioms and Meta-Metrics}
\label{sec:method:axioms}


\paragraph{Axiom 1: Reference Rationale Upper Bound}

In certain respects, the quality of $\mathbf{r}_i$ can be measured as $\rho(\mathbf{r}_i, \mathbf{\hat{y}}_i)$, the consistency between $\mathbf{r}_i$ and  $\mathbf{\hat{y}}_i$ (Sec. \ref{sec:background:eval}).
Intuitively, $\argmax_{\mathbf{r}_i} \rho(\mathbf{r}_i, \mathbf{\hat{y}}_i) = \mathbf{\hat{y}}_i$, since $\mathbf{\hat{y}}_i$ should be maximally consistent with itself.
It follows that $\Phi$ should yield a higher score for $\mathbf{\hat{y}}_i$ than for any other rationale, such that $\argmax_{\mathbf{r}_i} \Phi(\mathbf{r}_i) = \mathbf{\hat{y}}_i$.
We call $\mathbf{\hat{y}}_i$ the \textit{reference rationale}, which serves as an upper bound for $\Phi(\mathbf{r}_i)$.
Given the high complexity of NLM reasoning and the lack of constraints in free-text rationale generation (Sec. \ref{sec:background:gen}), it is difficult to establish any priors about $\rho(\mathbf{r}_i, \mathbf{\hat{y}}_i)$ for arbitrary $\mathbf{r}_i$.
To address this gap, the reference rationale upper bound provides a powerful invariance for analyzing free-text rationale quality.
Thus, \methodsp evaluates $\Phi$ by checking that $\Phi(\mathbf{\hat{y}}_i) > \Phi(\mathbf{r}_i)$ for various $\mathbf{r}_i$ (\eg $\mathbf{\hat{r}}_i$) that are likely to differ in meaning from $\mathbf{\hat{y}}_i$.

\paragraph{Axiom 2: Rationale Perturbation Sensitivity}
\label{sec:method:axiom2}

Rationale $\mathbf{r}_i$ should yield high $\rho(\mathbf{r}_i, \mathbf{\hat{y}}_i)$ if $\mathbf{r}_i$ contains the same semantic information as $\mathbf{\hat{y}}_i$, even if $\mathbf{r}_i$ and $\mathbf{\hat{y}}_i$ do not have the same surface form.
Conversely, $\mathbf{r}_i$ should yield low $\rho(\mathbf{r}_i, \mathbf{\hat{y}}_i)$ if $\mathbf{r}_i$ differs greatly in meaning from $\mathbf{\hat{y}}_i$.
Thus, $\Phi$ should be appropriately sensitive when either \textit{equivalent} (does not change $\mathbf{r}_i$'s meaning) or \textit{contrastive} (changes $\mathbf{r}_i$'s meaning) perturbations are applied to $\mathbf{r}_i$.
We denote the perturbed rationale as $\Psi(\mathbf{r}_i)$, where $\Psi$ is a perturbation function.
While $\Phi$ should be properly perturbation-sensitive for any rationale $\mathbf{r}_i$, we focus on perturbing $\mathbf{\hat{y}}_i$ in this work.
This is because the value of $\rho(\mathbf{r}_i, \mathbf{\hat{y}}_i)$ for arbitrary $\mathbf{r}_i$ is unclear, whereas we know $\rho(\mathbf{\hat{y}}_i, \mathbf{\hat{y}}_i)$ is high.
See Sec. \ref{sec:app:rationale_perturbation} for details about rationale perturbation.

\paragraph{Axiom 3: Robustness to Variation in $\mathcal{F}$}
\label{sec:method:axiom3}

In general, $\rho(\mathbf{r}_i, \mathbf{\hat{y}}_i)$ should be independent of whether $\mathbf{\hat{y}}_i = \mathbf{\dot{y}}_i$  (\ie $\mathcal{F}$'s task performance).
Since $\mathbf{\hat{y}}_i$ always provides full information about itself, it follows that $\Phi(\mathbf{\hat{y}}_i)$ should be roughly constant w.r.t. $\mathcal{F}$'s task performance.
Note that arbitrary $\mathbf{r}_i$ may not provide such an invariance.
For example, if we consider $\mathbf{\hat{r}}_i$ (\ie rationale outputted by $\mathcal{F}$), then $\mathbf{\hat{r}}_i$'s quality may depend on how $\mathcal{F}$ is trained.
Thus, in order to stably compare RLC metrics, we focus on analyzing RLC metrics w.r.t. $\mathbf{\hat{y}}_i$.
We vary $\mathcal{F}$'s performance in the following ways.

First, $\mathcal{F}$'s task performance can be changed by respectively varying the number of instances for training $\mathcal{F}$, number of noisy-labeled instances for training $\mathcal{F}$, and number of parameters in $\mathcal{F}$.
For each version of $\mathcal{F}$, we compute $\Phi(\mathbf{\hat{y}}_i)$.
A good $\Phi$ should yield similar $\Phi(\mathbf{\hat{y}}_i)$ across all $\mathcal{F}$ versions.

Second, we separately compute $\Phi(\mathbf{\hat{y}}_i)$ for the subpopulation of instances where $\mathcal{F}$ outputted correct predictions (\ie $\mathbf{\hat{y}}_i = \mathbf{\dot{y}}_i$) as well as the subpopulation of instances where $\mathcal{F}$ outputted incorrect predictions (\ie $\mathbf{\hat{y}}_i \neq \mathbf{\dot{y}}_i$).
A good $\Phi$ should yield similar $\Phi(\mathbf{\hat{y}}_i)$ across both subpopulations.





\begin{figure}[t!]
\centering
\includegraphics[width=\linewidth]{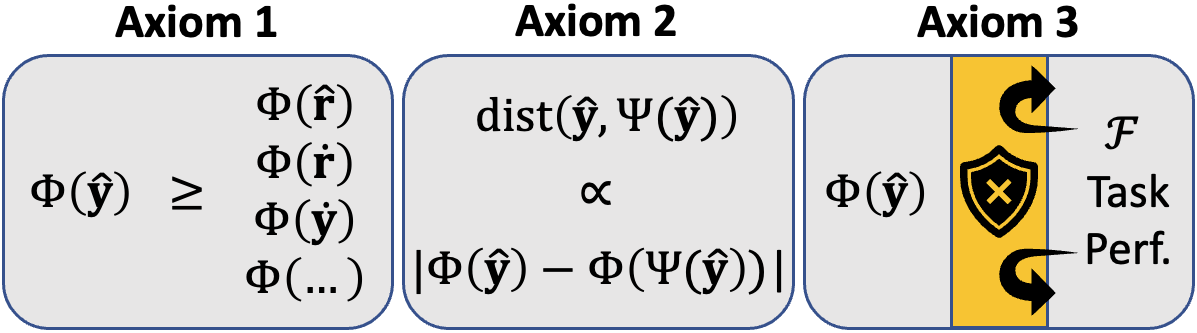}
\caption{\small \textbf{\methodsp Axioms.} FRAME evaluates RLC metrics ($\Phi$) based on three axioms: (1) good $\Phi$ should yield highest scores for reference rationales ($\mathbf{\hat{y}}_i$), which maximize RLC by construction; (2) good $\Phi$ should be appropriately sensitive to semantic perturbation of rationales ($\Psi(\mathbf{\hat{y}}_i)$); and (3) good $\Phi$ should be robust to variation in the the task LM's ($\mathcal{F}$) task performance.}
\label{fig:frame_axioms}
\vspace{-0.4cm}
\end{figure}

\subsection{\methodsp Implementation}
\label{sec:method:implementation}

\paragraph{Axiom 1: Reference Rationale Upper Bound}
For Axiom 1, we want to check if each RLC metric $\Phi$ yields the highest scores for $\mathbf{\hat{y}}_i$.
First, we report $\Phi(\mathbf{\hat{y}}_i)$, where higher scores are better.
Second, given a set of non-reference rationales $\mathbf{R}_{\text{NR}} = \{\mathbf{\hat{r}}_i$, $\dot{\mathbf{r}}_i$, $\dot{\mathbf{y}}_i\}$, we want to check if $\Phi$ yields higher scores for $\mathbf{\hat{y}}_i$ than for each $\mathbf{r}_{\text{NR}} \in \mathbf{R}_{\text{NR}}$.
To account for scaling differences across different $\Phi$, we want to compare rationales using $\Phi(\mathbf{\hat{y}}_i) / \Phi(\mathbf{r}_{\text{NR}})$ instead of $\Phi(\mathbf{\hat{y}}_i) - \Phi(\mathbf{r}_{\text{NR}})$.
However, $\Phi(\mathbf{r}_{\text{NR}})$ is negative if $\mathbbm{1}_{\mathcal{H}}(\mathbf{\hat{y}}_i | \mathbf{x}_i, \mathbf{r}_{\text{NR}}) < \mathbbm{1}_{\mathcal{G}}(\mathbf{\hat{y}}_i | \mathbf{x}_i)$, making $\Phi(\mathbf{\hat{y}}_i) / \Phi(\mathbf{r}_{\text{NR}})$ uninformative.
Thus, we instead compute accuracy ratio $\mathbbm{1}_{\mathcal{H}}(\mathbf{\hat{y}}_i | \mathbf{x}_i, \mathbf{\hat{y}}_i) / \mathbbm{1}_{\mathcal{H}}(\mathbf{\hat{y}}_i | \mathbf{x}_i, \mathbf{r}_{\text{NR}})$, since $\mathbbm{1}_{\mathcal{G}}(\mathbf{\hat{y}}_i | \mathbf{x}_i)$ is constant across all rationales.
To aggregate the accuracy ratios, we define the \textbf{M}ean \textbf{A}ccuracy \textbf{R}atio (\textbf{MAR}) as the mean $\mathbbm{1}_{\mathcal{H}}(\mathbf{\hat{y}}_i | \mathbf{x}_i, \mathbf{\hat{y}}_i) / \mathbbm{1}_{\mathcal{H}}(\mathbf{\hat{y}}_i | \mathbf{x}_i, \mathbf{r}_{\text{NR}})$ over all considered non-reference rationales.

\paragraph{Axiom 2: Rationale Perturbation Sensitivity}
For Axiom 2, we want to assess if each $\Phi$ is appropriately sensitive to semantic perturbation $\Psi$.
Given $\mathbf{\hat{y}}_i$ and $\Psi(\mathbf{\hat{y}}_i)$, we define the \textbf{A}bsolute \textbf{S}imulatability \textbf{D}ifference (\textbf{ASD}) as $\text{ASD}(a, b) = | \Phi(a) - \Phi(b) |$, with $a = \mathbf{\hat{y}}_i$ and $b = \Psi(\mathbf{\hat{y}}_i)$.
If $\Psi(\mathbf{\hat{y}}_i)$ is obtained via equivalent/contrastive perturbation, then lower/higher ASD is better, since $\mathbf{\hat{y}}_i$ and $\Psi(\mathbf{\hat{y}}_i)$ should be similar/dissimilar in meaning.

\paragraph{Axiom 3: Robustness to Variation in $\mathcal{F}$}
For Axiom 3, we want to assess whether each RLC metric $\Phi$ is robust to variation in: (A) the number of instances for training $\mathcal{F}$ (\ie 100\%, 50\%, 30\%, 10\%); (B) number of noisy-labeled instances for training $\mathcal{F}$ (\ie 0\%, 10\%, 30\%, 50\%); (C) number of parameters in $\mathcal{F}$ (\ie T5-Small, T5-Base, T5-Large); and (D) $\mathcal{F}$'s task performance (\ie $\hat{y}_i = \dot{y}_i$ \versus $\hat{y}_i \neq \dot{y}_i$).
For each variation factor (A)-(C), we compute $\Phi(\mathbf{\hat{y}}_i)$ based on each setting within the factor, then compute the mean $\mu(\Phi(\mathbf{\hat{y}}_i))$ and standard deviation $\sigma(\Phi(\mathbf{\hat{y}}_i))$ over all settings.
Generally, lower $\sigma(\Phi(\mathbf{\hat{y}}_i))$ indicates higher $\Phi$ robustness.
However, since $\mu(\Phi(\mathbf{\hat{y}}_i))$ can differ greatly across different $\Phi$, we want to account for such scaling differences when comparing $\sigma(\Phi(\mathbf{\hat{y}}_i))$.
Thus, we instead report the \textbf{S}im \textbf{C}oefficient of \textbf{V}ariation (\textbf{SCV}), defined as $\sigma(\Phi(\mathbf{\hat{y}}_i)) / \mu(\Phi(\mathbf{\hat{y}}_i))$.
Lower SCV is better, indicating higher $\Phi$ robustness to change in variation settings.
For the variation factor (D), we compare $\Phi(\mathbf{\hat{y}}_i)$ for $\mathcal{F}$'s correctly predicted test instances (\ie $\Phi_{\hat{y}_i = \dot{y}_i}(\mathbf{\hat{y}}_i)$) to $\Phi(\mathbf{\hat{y}}_i)$ for $\mathcal{F}$'s incorrectly predicted test instances (\ie $\Phi_{\hat{y}_i \neq \dot{y}_i}(\mathbf{\hat{y}}_i)$).
To do so, we report ASD$(a, b)$, with $a = \Phi_{\hat{y}_i = \dot{y}_i}(\mathbf{\hat{y}}_i)$ and $b = \Phi_{\hat{y}_i \neq \dot{y}_i}(\mathbf{\hat{y}}_i)$.

\begin{figure*}[t!]
\centering
\includegraphics[width=\textwidth]{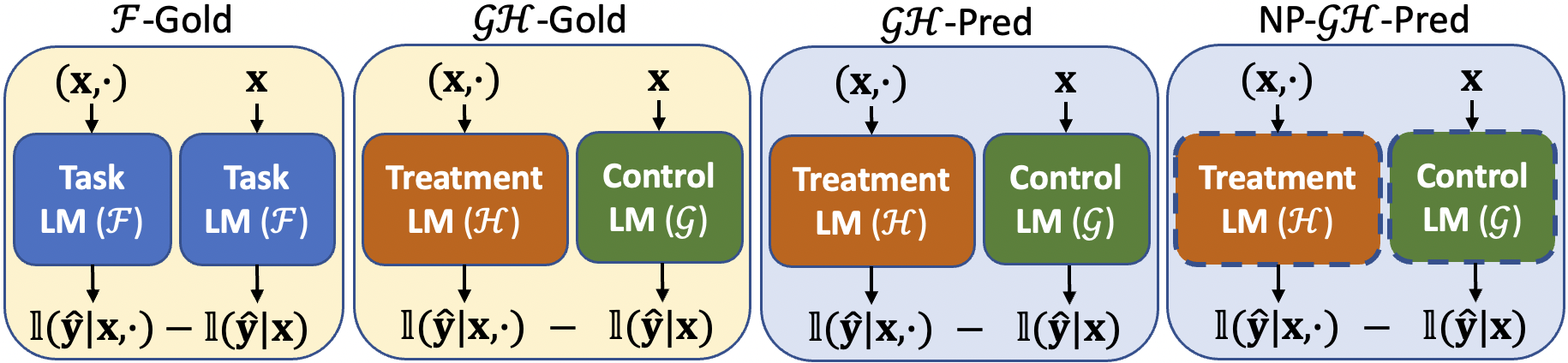}
\caption{\small \textbf{RLC Metrics.} We consider four RLC metrics (\textsc{$\mathcal{F}$-Gold, \textsc{$\mathcal{GH}$-Gold}, \textsc{$\mathcal{GH}$-Pred}}, \textsc{NP-$\mathcal{GH}$-Pred}) (Sec. \ref{sec:method:sim_metrics}). Metrics with ``$\mathcal{F}$'' use only $\mathcal{F}$ as the simulator. Metrics with ``$\mathcal{GH}$'' use $\mathcal{G}$ and $\mathcal{H}$ as simulators. Metrics with ``Gold'' (yellow background) involve training simulators to predict the gold label $\mathbf{\dot{y}}_i$. Metrics with ``Pred'' (blue background) involve training simulators to predict $\mathcal{F}$'s predicted label $\mathbf{\hat{y}}_i$. \textsc{NP-$\mathcal{GH}$-Pred} is the only metric without simulator pretraining (dotted lines), which can hurt RLC metrics' reliability (Sec. \ref{sec:experiments})}
\label{fig:sim_metrics}
\vspace{-0.4cm}
\end{figure*}


\subsection{RLC Metrics}
\label{sec:method:sim_metrics}

In this work, we consider four representative RLC metrics, which differ in whether $\mathcal{G}$ and $\mathcal{H}$ are: (A) separate from $\mathcal{F}$, (B) trained to predict $\mathbf{\hat{y}}_i$, and (C) pretrained on external corpora.
The first three (\textsc{$\mathcal{F}$-Gold, \textsc{$\mathcal{GH}$-Gold}, \textsc{$\mathcal{GH}$-Pred}}) are baselines, while the fourth (\textsc{NP-$\mathcal{GH}$-Pred}) is our proposed RLC metric (Fig. \ref{fig:sim_metrics}).
Note that prior free-text rationale evaluation works have also studied aspects like label leakage \cite{hase2020leakage}, input noise robustness \cite{wiegreffe2021measuring}, and feature importance agreement \cite{wiegreffe2021measuring}.
However, these aspects are orthogonal to the focus of our work (\ie (A)-(C)), as each existing RLC metric can still be categorized as one of the four we consider.
By default, we use T5-Base for $\mathcal{F}$, $\mathcal{G}$, and $\mathcal{H}$.
We describe each RLC metric below.

\paragraph{\textsc{$\mathcal{F}$-Gold}}
$\mathcal{F}$-Gold is a baseline RLC metric introduced by us.
Recall $\mathcal{F}$ is first pretrained on external corpora, then finetuned to predict $\mathbf{\dot{y}}_i$.
In \textsc{$\mathcal{F}$-Gold}, $\mathcal{G}$ is replaced by $\mathcal{F}$ with $\mathbf{x}_i$ as input, while $\mathcal{H}$ is replaced by $\mathcal{F}$ with $\mathbf{x}_i$ and $\mathbf{r}_i$ as input.
Thus, $\mathcal{G}$'s accuracy in predicting $\mathbf{\hat{y}}_i$ is trivially always 100\%, which limits $\mathcal{F}$-Gold's utility.
Unlike other metrics, \textsc{$\mathcal{F}$-Gold} does not need to assume $\mathcal{F}$, $\mathcal{G}$, and $\mathcal{H}$ are different LMs with similiar reasoning processes.

\paragraph{\textsc{$\mathcal{GH}$-Gold}}
In \textsc{$\mathcal{GH}$-Gold}, $\mathcal{G}$ and $\mathcal{H}$ are first pretrained on external corpora, then finetuned to predict $\mathbf{\dot{y}}_i$.
\textsc{$\mathcal{GH}$-Gold} is the RLC metric used in the evaluations proposed by \citet{wiegreffe2021measuring}.

\paragraph{\textsc{$\mathcal{GH}$-Pred}}
In \textsc{$\mathcal{GH}$-Pred}, $\mathcal{G}$ and $\mathcal{H}$ are first pretrained on external corpora, then finetuned to predict $\mathbf{\hat{y}}_i$. 
\textsc{$\mathcal{GH}$-Pred} is very similar in spirit to the RLC metric used in \citet{hase2020leakage} (Sec. \ref{sec:background}).

\paragraph{\textsc{NP-$\mathcal{GH}$-Pred}}
\textsc{NP-$\mathcal{GH}$-Pred} is our proposed RLC metric.
Unlike the other metrics, $\mathcal{G}$ and $\mathcal{H}$ are not pretrained on external corpora.
Here, $\mathcal{G}$ and $\mathcal{H}$ are randomly initialized, then trained to predict $\mathbf{\hat{y}}_i$.
We hypothesize that this will enable us to better isolate the rationale's contributions from the pretraining knowledge obtained by pretrained versions of $\mathcal{G}$ and $\mathcal{H}$.
However, not pretraining $\mathcal{G}$ and $\mathcal{H}$ may weaken the assumption that $\mathcal{F}$, $\mathcal{G}$, and $\mathcal{H}$ have similiar reasoning processes.

\section{Experiments} 
\label{sec:experiments}


In this section, we present empirical results for \methodsp.
First, for LM simulators, we investigate \methodsp Axioms 1-3 in the context of (automatic) faithfulness evaluation, demonstrating the relative effectiveness of the \textsc{NP-$\mathcal{GH}$-Pred} RLC metric in measuring rationale quality (Sec. \ref{sec:exp:faith}).
Second, for human simulators, we conduct a user study about Axiom 1 to show \method's applicability to (manual) plausibility evaluation, while further establishing \textsc{NP-$\mathcal{GH}$-Pred}'s effectiveness (Sec. \ref{sec:exp:plaus}).
For all results, we report the mean and standard deviation over three seeds.
In all tables, best performance is highlighted in \colorbox{lgreen}{green}, while second-best performance (if applicable) is highlighted in \colorbox{lblue}{blue}.

\paragraph{Normalized Relative Gain (NRG)}
For each axiom, we would like to summarize all metrics as a single aggregate metric, but this may not be straightforward if different metrics have different scales or different optimal values (\eg higher is better \versus lower is better).
Thus, for each axiom, we summarize all metrics using the Normalized Relative Gain (NRG) metric \citep{chan2022unirex}, which normalizes all metrics' scores to values in $[0, 1]$ then outputs the final score by computing the mean of the normalized scores.

\subsection{Datasets}
We consider closed-set and multi-choice $m$-class text classification tasks.
In closed-set classification (\eg natural language inference (NLI)), the same $m$-class label space is used for all instances.
In multi-choice classification (\eg commonsense QA), each instance has a different set of $m$ choices.
Following prior works \citep{hase2020leakage, wiegreffe2021measuring}, we experiment with the e-SNLI \citep{camburu2018snli}, CoS-E v1.0 \citep{rajani2019explain}, and CoS-E v1.1 \citep{rajani2019explain} datasets.
e-SNLI is an NLI dataset, while CoS-E v1.0 and CoS-E v1.1 are commonsense QA datasets.

\subsection{LM Simulators}
\label{sec:exp:faith}

\paragraph{Axiom 1: Reference Rationale Upper Bound}

Table \ref{tab:axiom1} displays results for Axiom 1.
\textsc{NP-$\mathcal{GH}$-Pred} performs best overall, achieving the highest $\Phi(\mathbf{\hat{y}}_i)$, MAR, and NRG on all datasets, besides having the second-highest MAR on e-SNLI.
None of the other metrics perform consistently well across datasets.
For mean NRG over all datasets, \textsc{NP-$\mathcal{GH}$-Pred} beats the strongest per-dataset baseline by 41.7\%, hence best satisfying Axiom 1.
This suggests that simulator pretraining worsens RLC metrics' ability to capture relative rationale-label association across different rationales.

\begin{table}[ht!]
\centering
\scalebox{0.6}{
\begin{tabular}{ccccc}
    \toprule
    \multirow{1}{*}{\textbf{Dataset}} & \multirow{1}{*}{\textbf{RLC Metric}} & $\Phi(\mathbf{\hat{y}}_i)$ ($\uparrow$) & MAR ($\uparrow$) & NRG ($\uparrow$) \\
    \midrule
    \multirow{4}{*}{e-SNLI} & \textsc{$\mathcal{F}$-Gold} & -7.27~($\pm$0.57) & \cellcolor{lgreen}{1.45~($\pm$0.01)} & \cellcolor{lblue}{0.50} \\
    & \textsc{$\mathcal{GH}$-Gold} & -1.70~($\pm$8.25) & 1.01~($\pm$0.07) & 0.04 \\
    & \textsc{$\mathcal{GH}$-Pred} & \cellcolor{lblue}{9.10~($\pm$0.61)} & 1.01~($\pm$0.01)& 0.13 \\
    & \textsc{NP-$\mathcal{GH}$-Pred} & \cellcolor{lgreen}{54.77~($\pm$4.27)} & \cellcolor{lblue}{1.15~($\pm$0.01)} & \cellcolor{lgreen}{0.66} \\
    
    \midrule
    
    \multirow{4}{*}{CoS-E v1.0} & \textsc{$\mathcal{F}$-Gold} & -34.39~($\pm$1.83) & 0.85~($\pm$0.01) & 0.00 \\
    & \textsc{$\mathcal{GH}$-Gold} & 17.93~($\pm$1.81) & \cellcolor{lblue}{1.33~($\pm$0.02)} & 0.65 \\
    & \textsc{$\mathcal{GH}$-Pred} & \cellcolor{lblue}{18.60~($\pm$0.86)} & \cellcolor{lblue}{1.33~($\pm$0.04)} & \cellcolor{lblue}{0.66} \\
    & \textsc{NP-$\mathcal{GH}$-Pred} & \cellcolor{lgreen}{23.83~($\pm$2.17)} & \cellcolor{lgreen}{2.04~($\pm$0.09)} & \cellcolor{lgreen}{1.00} \\
    
    \midrule
    
    \multirow{4}{*}{CoS-E v1.11} & \textsc{$\mathcal{F}$-Gold} & -35.63~($\pm$1.63) & 0.83~($\pm$0.02) & 0.00 \\
    & \textsc{$\mathcal{GH}$-Gold} & 21.16~($\pm$2.36) & \cellcolor{lblue}{1.47~($\pm$0.06)} & 0.70 \\
    & \textsc{$\mathcal{GH}$-Pred} & \cellcolor{lblue}{22.17~($\pm$1.04)} & 1.46~($\pm$0.03) & \cellcolor{lblue}{0.71} \\
    & \textsc{NP-$\mathcal{GH}$-Pred} & \cellcolor{lgreen}{23.48~($\pm$0.60)} & \cellcolor{lgreen}{2.26~($\pm$0.09)} & \cellcolor{lgreen}{1.00} \\
    
    \bottomrule 
\end{tabular}
}
\caption{\small \textbf{Axiom 1 (LM Simulators).} We evaluate RLC metrics' ability to use reference rationale as a faithfulness upper bound. \textsc{NP-$\mathcal{GH}$-Pred} performs best overall, while none of the RLC metrics are consistently good across all three datasets.
}
\label{tab:axiom1}
\end{table}

\paragraph{Axiom 2: Rationale Perturbation Sensitivity}

Table \ref{tab:axiom2} displays results for Axiom 2.
\textsc{NP-$\mathcal{GH}$-Pred} performs best on e-SNLI (w.r.t. Contrastive ASD and NRG) and competitively on CoS-E v1.0 and CoS-E v1.11.
However, \textsc{$\mathcal{GH}$-Gold} performs best overall, yielding the second-highest NRG on e-SNLI and the highest NRG on CoS-E v1.0 and CoS-E v1.11.
We find that the equivalent perturbation property is easier to satisfy, as all metrics except \textsc{$\mathcal{F}$-Gold} yield low Equivalent ASD.
Meanwhile, the contrastive perturbation property is harder to satisfy, as only \textsc{$\mathcal{GH}$-Gold} consistently achieves high Contrastive ASD.
These results confirm the limitation of using $\mathcal{F}$ as the simulator, while suggesting that using $\mathbf{\dot{y}}_i$ as simulator supervision may  help RLC metrics' semantic sensitivity.    

\begin{table}[ht!]
\centering
\scalebox{0.6}{
\begin{tabular}{ccccc}
    \toprule
    \multirow{2}{*}{\textbf{Dataset}} & \multirow{2}{*}{\textbf{RLC Metric}} & \multicolumn{1}{c}{\textbf{Equivalent}} & \multicolumn{1}{c}{\textbf{Contrastive}} & \textbf{All} \\
    \cmidrule(lr){3-3} \cmidrule(lr){4-4} \cmidrule(lr){5-5}
    & & ASD ($\downarrow$) & ASD ($\uparrow$) & NRG ($\uparrow$) \\
    \midrule
    \multirow{4}{*}{e-SNLI} & \textsc{$\mathcal{F}$-Gold} & 11.80~($\pm$0.60) & 12.50~($\pm$0.44) & 0.10 \\
    & \textsc{$\mathcal{GH}$-Gold} & 6.10~($\pm$7.96) & \cellcolor{lblue}{40.53~($\pm$7.03)} & \cellcolor{lblue}{0.74} \\
    & \textsc{$\mathcal{GH}$-Pred} & \cellcolor{lgreen}{0.00~($\pm$0.00)} & 5.23~($\pm$1.02) & 0.50 \\
    & \textsc{NP-$\mathcal{GH}$-Pred} & \cellcolor{lblue}{0.07~($\pm$0.12)} & \cellcolor{lgreen}{40.83~($\pm$2.63)} & \cellcolor{lgreen}{1.00} \\
    
    \midrule
    
    \multirow{4}{*}{CoS-E v1.0} & \textsc{$\mathcal{F}$-Gold} & 5.61~($\pm$2.00) & \cellcolor{lblue}{26.91~($\pm$2.37)} & 0.09 \\
    & \textsc{$\mathcal{GH}$-Gold} & \cellcolor{lblue}{0.67~($\pm$0.40)} & \cellcolor{lgreen}{83.09~($\pm$13.16)} & \cellcolor{lgreen}{0.94} \\
    & \textsc{$\mathcal{GH}$-Pred} & \cellcolor{lgreen}{0.00~($\pm$0.00)} & 15.09~($\pm$4.29)) & 0.50 \\
    & \textsc{NP-$\mathcal{GH}$-Pred} & 0.74~($\pm$0.76) & 26.21~($\pm$2.79) & \cellcolor{lblue}{0.52} \\
    
    \midrule
    
    \multirow{4}{*}{CoS-E v1.11} & \textsc{$\mathcal{F}$-Gold} & 4.04~($\pm$0.31) & \cellcolor{lblue}{26.13~($\pm$2.15)} & 0.04 \\
    & \textsc{$\mathcal{GH}$-Gold} & \cellcolor{lblue}{0.96~($\pm$1.65)} & \cellcolor{lgreen}{84.14~($\pm$21.59)} & \cellcolor{lgreen}{0.93} \\
    & \textsc{$\mathcal{GH}$-Pred} & \cellcolor{lgreen}{0.00~($\pm$0.00)} & 21.57~($\pm$1.04)) & \cellcolor{lblue}{0.50} \\
    & \textsc{NP-$\mathcal{GH}$-Pred} & 1.42~($\pm$0.48) & 25.55~($\pm$0.99) & 0.36 \\
    
    \bottomrule 
\end{tabular}
}
\vspace{0.3cm}
\caption{\small \textbf{Axiom 2 (LM Simulators).} We evaluate RLC metrics' sensitivity to semantic perturbation of rationales. We find that \textsc{$\mathcal{GH}$-Gold} performs best overall, while \textsc{NP-$\mathcal{GH}$-Pred} also consistently performs well across all three datasets.
}
\label{tab:axiom2}
\end{table}

\paragraph{Axiom 3: Robustness to Variation in $\mathcal{F}$}

\begin{figure*}[h!]
	\begin{subfigure}[b]{0.33\linewidth}
		\centering
		\includegraphics[width=\textwidth]{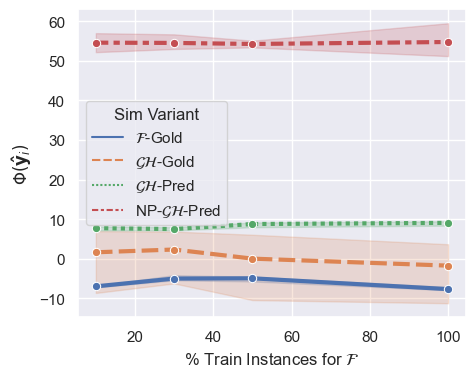}
		\caption{e-SNLI}
		\label{fig:pct_instances:esnli}
	\end{subfigure}\hfill
	\begin{subfigure}[b]{0.33\linewidth}
		\centering
		\includegraphics[width=\textwidth]{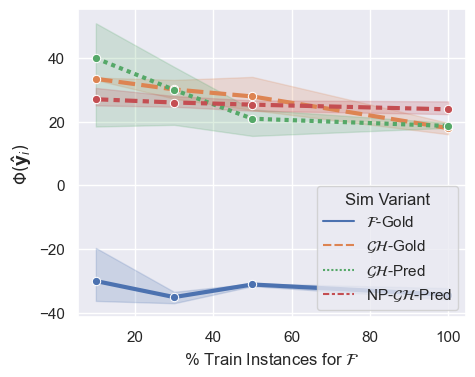}
		\caption{CoS-E v1.0}
		\label{fig:pct_instances:cose_v1-0}
	\end{subfigure}\hfill
	\begin{subfigure}[b]{0.33\linewidth}
		\centering
		\includegraphics[width=\textwidth]{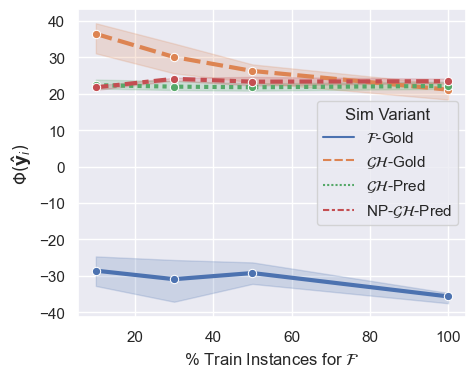}
		\caption{CoS-E v1.11}
		\label{fig:pct_instances:cose_v1-11}
	\end{subfigure}
	\caption{\small \textbf{Axiom 3 (LM Simulators): \% Train Instances for $\mathcal{F}$ \versus $\Phi(\mathbf{\hat{y}}_i)$.} Flatter curves mean greater robustness to change in $\mathcal{F}$ task performance, by varying the percentage of train instances for $\mathcal{F}$.}
	\label{fig:pct_instances}
\end{figure*}

\begin{figure*}[h!]
	\begin{subfigure}[b]{0.33\linewidth}
		\centering
		\includegraphics[width=\textwidth]{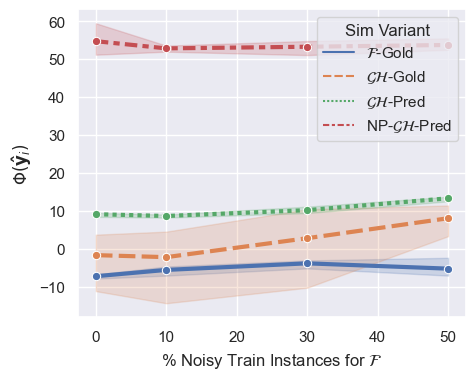}
		\caption{e-SNLI}
		\label{fig:pct_noisy:esnli}
	\end{subfigure}\hfill
	\begin{subfigure}[b]{0.33\linewidth}
		\centering
		\includegraphics[width=\textwidth]{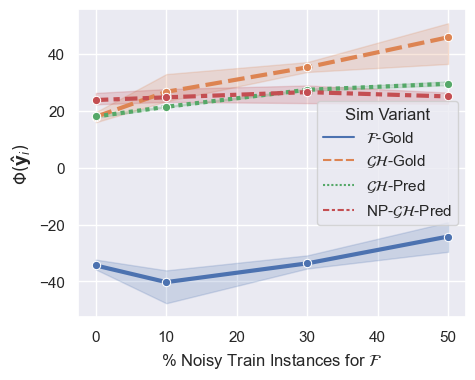}
		\caption{CoS-E v1.0}
		\label{fig:pct_noisy:cose_v1-0}
	\end{subfigure}\hfill
	\begin{subfigure}[b]{0.33\linewidth}
		\centering
		\includegraphics[width=\textwidth]{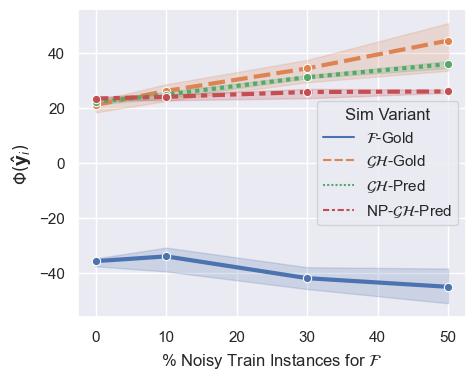}
		\caption{CoS-E v1.11}
		\label{fig:pct_noisy:cose_v1-11}
	\end{subfigure}
	\caption{\small \textbf{Axiom 3 (LM Simulators): \% Noisy Train Instances for $\mathcal{F}$ \versus $\Phi(\mathbf{\hat{y}}_i)$.} Flatter curves mean greater robustness to change in $\mathcal{F}$ task performance, by varying the percentage of noisy train instances for $\mathcal{F}$.}
	\label{fig:pct_noisy}
\end{figure*}

\begin{figure*}[h!]
	\begin{subfigure}[b]{0.33\linewidth}
		\centering
		\includegraphics[width=\textwidth]{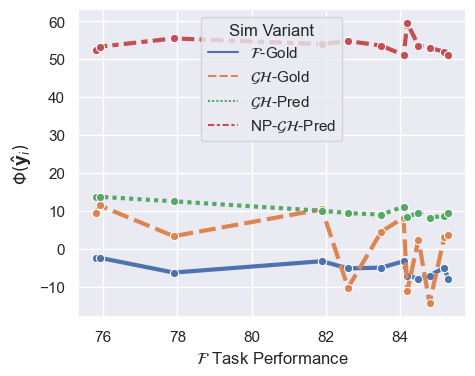}
		\caption{e-SNLI}
		\label{fig:perf_vs_sim:esnli}
	\end{subfigure}\hfill
	\begin{subfigure}[b]{0.33\linewidth}
		\centering
		\includegraphics[width=\textwidth]{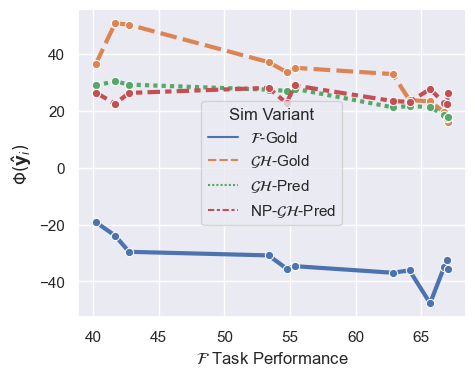}
		\caption{CoS-E v1.0}
		\label{fig:perf_vs_sim:cose_v1-0}
	\end{subfigure}\hfill
	\begin{subfigure}[b]{0.33\linewidth}
		\centering
		\includegraphics[width=\textwidth]{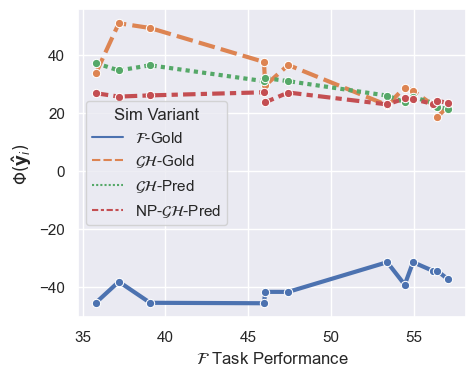}
		\caption{CoS-E v1.11}
		\label{fig:perf_vs_sim:cose_v1-11}
	\end{subfigure}
	\caption{\small \textbf{Axiom 3 (LM Simulators): $\mathcal{F}$ Task Performance \versus $\Phi(\mathbf{\hat{y}}_i)$.} Flatter curves mean greater robustness to change in $\mathcal{F}$ task performance, by varying the percentage of noisy train instances for $\mathcal{F}$. Unlike in Fig. \ref{fig:pct_noisy}, we directly plot $\Phi(\mathbf{\hat{y}}_i)$ as a function of $\mathcal{F}$ task performance here.}
	\label{fig:perf_vs_sim}
\end{figure*}

Table \ref{tab:axiom3} and Fig. \ref{fig:pct_instances}-\ref{fig:perf_vs_sim} display results for Axiom 3.
In Table \ref{tab:axiom3}, \textsc{NP-$\mathcal{GH}$-Pred} greatly outperforms all baselines all meta-metrics, yielding a perfect NRG of 1.00 on all datasets.
None of the other metrics perform consistently well across datasets.
Similarly, in Fig. \ref{fig:pct_instances}-\ref{fig:perf_vs_sim}, \textsc{NP-$\mathcal{GH}$-Pred} yields much flatter curves than other RLC metrics do, indicating its relative robustness to changes in $\mathcal{F}$'s number of train instances (Fig. \ref{fig:pct_instances}), number of noisy train instances (Fig. \ref{fig:pct_noisy}), and task performance (Fig. \ref{fig:perf_vs_sim}).
For mean NRG over all datasets, \textsc{NP-$\mathcal{GH}$-Pred} beats the strongest per-dataset baselines by 42.9\%, hence best satisfying Axiom 3.
This suggests that simulator pretraining hurts RLC metrics' ability to disentangle rationale-label association from $\mathcal{F}$'s task performance.

\begin{table}[h]
\centering
\scalebox{0.44}{
\begin{tabular}{ccccccc}
    \toprule
    \multirow{2}{*}{\textbf{Dataset}} & \multirow{2}{*}{\textbf{RLC Metric}} & \multicolumn{1}{c}{\textbf{\% Train}} & \multicolumn{1}{c}{\textbf{\% Noisy Train}} & \multicolumn{1}{c}{\textbf{Capacity}} & \multicolumn{1}{c}{\textbf{Subpop.}} & \textbf{All} \\
    \cmidrule(lr){3-3} \cmidrule(lr){4-4} \cmidrule(lr){5-5} \cmidrule(lr){6-6} \cmidrule(lr){7-7}
    & & SCV ($\downarrow$) & SCV ($\downarrow$) & SCV ($\downarrow$) & ASD ($\downarrow$) & NRG ($\uparrow$) \\
    \midrule
    \multirow{4}{*}{e-SNLI} & \textsc{$\mathcal{F}$-Gold} & 0.27~($\pm$0.06) & 0.34~($\pm$0.13) & 0.81~($\pm$0.18) & \cellcolor{lblue}{6.81~($\pm$1.51)} & 0.27 \\
    & \textsc{$\mathcal{GH}$-Gold} & 0.14~($\pm$0.26) & 0.71~($\pm$0.26) & 0.65~($\pm$0.21) & 20.18~($\pm$4.20) & 0.26 \\
    & \textsc{$\mathcal{GH}$-Pred} & \cellcolor{lblue}{0.11~($\pm$0.02)} & \cellcolor{lblue}{0.21~($\pm$0.02)} & \cellcolor{lblue}{0.16~($\pm$0.04)} & 32.97~($\pm$3.36) & \cellcolor{lblue}{0.69} \\
    & \textsc{NP-$\mathcal{GH}$-Pred} & \cellcolor{lgreen}{0.02~($\pm$0.01)} & \cellcolor{lgreen}{0.03~($\pm$0.02)} & \cellcolor{lgreen}{0.04~($\pm$0.00)} & \cellcolor{lgreen}{1.85~($\pm$1.12)} & \cellcolor{lgreen}{1.00} \\
    
    \midrule
    
    \multirow{4}{*}{CoS-E v1.0} & \textsc{$\mathcal{F}$-Gold} & \cellcolor{lblue}{0.13~($\pm$0.11)} & \cellcolor{lblue}{0.21~($\pm$0.12)} & 0.55~($\pm$0.05) & 23.41~($\pm$1.79) & \cellcolor{lblue}{0.71} \\
    & \textsc{$\mathcal{GH}$-Gold} & 0.25~($\pm$0.13) & 0.40~($\pm$0.22) & 0.43~($\pm$0.18) & 27.56~($\pm$0.87) & 0.15 \\
    & \textsc{$\mathcal{GH}$-Pred} & 0.32~($\pm$0.19) & 0.22~($\pm$0.04) & \cellcolor{lblue}{0.08~($\pm$0.04)} & \cellcolor{lblue}{23.20~($\pm$4.39)} & 0.30 \\
    & \textsc{NP-$\mathcal{GH}$-Pred} & \cellcolor{lgreen}{0.08~($\pm$0.02)} & \cellcolor{lgreen}{0.10~($\pm$0.00)} & \cellcolor{lgreen}{0.05~($\pm$0.02)} & \cellcolor{lgreen}{5.37~($\pm$4.51)} & \cellcolor{lgreen}{1.00} \\
    
    \midrule
    
    \multirow{4}{*}{CoS-E v1.11} & \textsc{$\mathcal{F}$-Gold} & \cellcolor{lblue}{0.14~($\pm$0.03)} & \cellcolor{lblue}{0.16~($\pm$0.07)} & 0.63~($\pm$0.08) & 25.48~($\pm$1.77) & 0.57 \\
    & \textsc{$\mathcal{GH}$-Gold} & 0.23~($\pm$0.05) & 0.32~($\pm$0.09) & 0.40~($\pm$0.05) & \cellcolor{lblue}{22.61~($\pm$1.87)} & 0.00 \\
    & \textsc{$\mathcal{GH}$-Pred} & \cellcolor{lgreen}{0.05~($\pm$0.02)} & 0.22~($\pm$0.02) & \cellcolor{lblue}{0.32~($\pm$0.01)} & 25.47~($\pm$1.22) & \cellcolor{lblue}{0.70} \\
    & \textsc{NP-$\mathcal{GH}$-Pred} & \cellcolor{lgreen}{0.05~($\pm$0.01)} & \cellcolor{lgreen}{0.07~($\pm$0.01)} & \cellcolor{lgreen}{0.09~($\pm$0.02)} & \cellcolor{lgreen}{6.28~($\pm$2.05)} & \cellcolor{lgreen}{1.00} \\
    
    \bottomrule 
\end{tabular}
}
\caption{\small \textbf{Axiom 3 (LM Simulators).} We evaluate RLC metrics' robustness to four types of variation in $\mathcal{F}$. Here, \textsc{NP-$\mathcal{GH}$-Pred} greatly outperforms other RLC metrics across all datasets.
}
\label{tab:axiom3}
\end{table}



\begin{table*}[t]
\centering
\scalebox{0.7}{
\begin{tabular}{cccccccccc}
    \toprule
    \multirow{2}{*}{\textbf{RLC Metric}} & \multicolumn{2}{c}{\textbf{Sim-Based Scores}} & \multicolumn{5}{c}{\textbf{Confidence}} \\
    \cmidrule(lr){2-3} \cmidrule(lr){4-8}
    
    & $\Phi(\mathbf{\hat{y}}_i)$ ($\uparrow$) & MAR ($\uparrow$) & $\mathbbm{1}_{\mathcal{G}}(\mathbf{x})$ ($\downarrow$) & $\mathbbm{1}_{\mathcal{H}}(\mathbf{x}, \mathbf{\hat{r}})$ ($\downarrow$) & $\mathbbm{1}_{\mathcal{H}}(\mathbf{x}, \mathbf{\dot{r}})$ ($\downarrow$) & $\mathbbm{1}_{\mathcal{H}}(\mathbf{x}, \mathbf{\dot{y}})$ ($\downarrow$) & $\mathbbm{1}_{\mathcal{H}}(\mathbf{x}, \mathbf{\hat{y}})$ ($\uparrow$)  \\

    \midrule
    \textsc{$\mathcal{GH}$-Gold} & 51.60~($\pm$28.00) & 2.03~($\pm$0.53) & 2.94~($\pm$0.37) & 3.06~($\pm$0.47) & 3.47~($\pm$0.35) & 3.72~($\pm$0.36) & 3.41~($\pm$0.66) \\
    
    \textsc{NP-$\mathcal{GH}$-Pred} & \cellcolor{lgreen}{71.60~($\pm$7.65)} & \cellcolor{lgreen}{3.18~($\pm$0.60)} & \cellcolor{lgreen}{1.13~($\pm$0.25)} & \cellcolor{lgreen}{1.82~($\pm$0.66)} & \cellcolor{lgreen}{1.76~($\pm$0.73)} & \cellcolor{lgreen}{2.84~($\pm$1.16)} & \cellcolor{lgreen}{3.44~($\pm$0.86)} \\
    
    \bottomrule 
\end{tabular}
}
\caption{\small \textbf{Axiom 1 (Human Simulators).} We conduct a user study to investigate \method's applicability to plausibility evaluation (\ie using human simulators) w.r.t. Axiom 1. Our results mirror those for faithfulness evaluation, with \textsc{NP-$\mathcal{GH}$-Pred} consistently outperforming \textsc{$\mathcal{GH}$-Gold}.
}
\label{tab:plaus}
\end{table*}

\subsection{Human Simulators}
\label{sec:exp:plaus}

Beyond faithfulness (Sec. \ref{sec:exp:faith}), we want to show that \methodsp can also be used to evaluate rationale plausibility.
Whereas faithfulness evaluation can be automated due to its use of LM simulators, plausibility evaluation requires significant manual effort from human simulators.
Since Axioms 2-3 have many more experiment settings than Axiom 1, they require much more manual effort for plausibility evaluation.
Thus, as a proof of concept, we conduct a plausibility user study based on Axiom 1, using a subset of the four RLC metrics.

Our goal is to verify that our Axiom 1 conclusions for faithfulness (Table \ref{tab:axiom1}) also hold for plausibility.
This means we must reframe the RLC metrics w.r.t. human simulators, which requires three more assumptions.
First, we assume human simulators are pretrained, since they should possess general language understanding abilities and world knowledge.
Second, we assume human simulators already know how to solve the given task (\ie already finetuned to predict $\mathbf{\dot{y}}_i$), using either $\mathbf{x}_i$ or $(\mathbf{x}_i, \mathbf{r}_i$) as input.
Third, we assume human simulators can learn patterns from few training examples.

Based on these assumptions, we compare \textsc{$\mathcal{GH}$-Gold} and \textsc{NP-$\mathcal{GH}$-Pred}.
For \textsc{$\mathcal{GH}$-Gold}, given the assumed pretraining/finetuning, we simply compute $\Phi$ without further training.
However, for \textsc{NP-$\mathcal{GH}$-Pred}, we need to disable pretraining and train the simulator to predict $\mathbf{\hat{y}}_i$.
To simulate the disabling of pretraining, we encrypt every character in the dataset using a Caesar cipher \cite{savarese1999caesar} with right shift 1 (\eg \textit{hello} $\rightarrow$ \textit{ifmmp}).
Each encrypted instance still encodes the same information as before, but encryption hinders human simulators from leveraging their language priors.
Thus, human simulators must rely on patterns learned while they are newly trained to predict $\mathbf{\hat{y}}_i$.

We uniformly sample 50 test instances from CoS-E v1.0, then ask five human annotators to serve as simulators for Axiom 1, as both \textsc{$\mathcal{GH}$-Gold} and \textsc{NP-$\mathcal{GH}$-Pred}.
Besides computing $\Phi(\mathbf{\hat{y}}_i)$ and MAR, we also ask annotators to rate their confidence for each sim accuracy term (\ie $\mathbbm{1}_{\mathcal{G}}(\mathbf{\hat{y}}_i | \mathbf{x}_i)$, $\mathbbm{1}_{\mathcal{H}}(\mathbf{\hat{y}}_i | \mathbf{x}_i, \mathbf{\hat{r}}_i)$, $\mathbbm{1}_{\mathcal{H}}(\mathbf{\hat{y}}_i | \mathbf{x}_i, \mathbf{\dot{r}}_i)$, $\mathbbm{1}_{\mathcal{H}}(\mathbf{\hat{y}}_i | \mathbf{x}_i, \mathbf{\dot{y}}_i)$, $\mathbbm{1}_{\mathcal{H}}(\mathbf{\hat{y}}_i | \mathbf{x}_i, \mathbf{\hat{y}}_i)$), on a 4-point Likert scale (\ie higher rating $\rightarrow$ higher confidence).

Table \ref{tab:plaus} displays the user study results.
On $\Phi(\mathbf{\hat{y}}_i)$ and MAR, \textsc{NP-$\mathcal{GH}$-Pred} greatly outperforms \textsc{$\mathcal{GH}$-Gold}, corroborating the results in Table \ref{tab:axiom1}.
Specifically, across $\Phi(\mathbf{\hat{y}}_i)$ and MAR, \textsc{NP-$\mathcal{GH}$-Pred} yields a 47.7\% mean improvement over \textsc{$\mathcal{GH}$-Gold}, while also having much lower standard deviation on $\Phi(\mathbf{\hat{y}}_i)$. 
Furthermore, \textsc{NP-$\mathcal{GH}$-Pred} yields much higher confidence than \textsc{$\mathcal{GH}$-Gold} for $\mathbbm{1}_{\mathcal{H}}(\mathbf{\hat{y}}_i | \mathbf{x}, \mathbf{\hat{y}})$, relative to the other four settings.
In other words, for \textsc{NP-$\mathcal{GH}$-Pred}, annotators not only yield higher $\Phi(\mathbf{\hat{y}})$, but also do so more confidently.
These results demonstrate \method's applicability to plausibility evaluation as well as \textsc{NP-$\mathcal{GH}$-Pred}'s effectiveness as a plausibility metric.

\section{Related Work} 
\label{sec:related_work}

\paragraph{Extractive Rationale Evaluation}


Most rationale evaluation works focus on extractive rationales.
Existing faithfulness metrics for extractive rationales are based on rationale-label association, either via heuristic functions \citep{deyoung2019eraser, shrikumar2017learning, sanyal2021discretized, hase2020evaluating} or model retraining \citep{hooker2018benchmark, pruthi2020evaluating}.
For example, \citet{pruthi2020evaluating} measures the faithfulness of a teacher LM's rationales as how well a student LM can predict the teacher's predicted labels when regularized using those rationales.
Meanwhile, existing plausibility metrics for extractive rationales are based on rationale-label association \citep{hase2020evaluating, strout2019human, chan2022unirex} or gold rationale similarity \citep{deyoung2019eraser, chan2022unirex}.
However, evaluating plausibility via gold rationale similarity is noisy because gold rationales are defined w.r.t. the gold label, not the task LM's predicted label \cite{chan2022unirex}.
 
Like \citet{pruthi2020evaluating}, our \textsc{NP-$\mathcal{GH}$-Pred} metric trains a student  (simulator) LM to predict a teacher (task) LM's labels, though we focus on free-text rationales and simply append the rationale to the student's input.
We do not apply \methodsp to extractive rationales, since it is not straightforward to construct extractive reference rationales.

\paragraph{Free-Text Rationale Evaluation}

Few works consider free-text rationale evaluation.
Like for extractive rationales, free-text rationale faithfulness is evaluated via rationale-label association, while  
Existing faithfulness metrics for free-text rationales are based on rationale-label association and implemented via sim (Sec. \ref{sec:background:eval}) \cite{hase2020leakage, wiegreffe2021measuring}, which evaluates rationales w.r.t. its predictiveness of the task LM's labels \cite{doshi2017towards}.
\citet{hase2020leakage} proposes LAS, a sim-based metric that controls for label leakage by the rationale. 
\citet{wiegreffe2021measuring} proposes two sim-based metrics: RE measures sim w.r.t. input noise robustness, while FIA measures sim w.r.t. feature importance agreement.
Existing plausibility metrics for free-text rationales can be based on sim \cite{hase2020leakage} or gold rationale similarity \cite{narang2020wt5}.

Crucially, existing RLC metrics pretrain the simulator LM on external corpora, which makes it difficult to isolate the rationale's contribution from the simulator LM's pretraining knowledge.
Using \method, we show this results in oversensitivity to the task LM's task performance and that our non-pretrained \textsc{NP-$\mathcal{GH}$-Pred} performs better on \methodsp meta-metrics.
Our contributions are orthogonal to those in LAS, RE, and FIA, as \textsc{NP-$\mathcal{GH}$-Pred} can be applied to all of these metrics.



\section{Conclusion} 
\label{sec:conclusion}

In this paper, we showed that using non-pretrained simulators (\ie \textsc{NP-$\mathcal{GH}$-Pred}) for RLC-based rationale evaluation can better satisfy the desired \methodsp properties than using pretrained simulators.
While these results demonstrate the relative effectiveness of the \textsc{NP-$\mathcal{GH}$-Pred} RLC metric, they also reveal the inherent limitations of using sim to evaluate rationale quality.
First, although reference rationales maximize RLC, they generally provide very limited information about an LM's or human's reasoning process.
This suggests that RLC alone cannot sufficiently capture faithfulness or plausibility.
Second, sim requires that the task LM and simulator are distinct, or else the simulator will trivially achieve perfect accuracy when using the original task input.
As a result, sim-based faithfulness evaluation calls for the strong assumption that the simulator's reasoning process is representative of the task LM's.
As RLC metrics can vary w.r.t. how the simulator is trained, it is unclear which simulator training procedures are acceptable for justifying the above assumption.
Though RLC can provide useful signal about rationale quality, these limitations suggest that future work on rationale evaluation should move beyond RLC.

\bibliography{references}
\bibliographystyle{acl_natbib}

\newpage
\appendix
\section{Appendix} 
\label{sec:appendix}



\subsection{Rationale Perturbation}
\label{sec:app:rationale_perturbation}


\paragraph{Equivalent Perturbation}
Let $\Psi_e$ be an equivalent perturbation function which modifies $\mathbf{r}_i$'s surface form without changing $\mathbf{r}_i$'s semantic content.
That is, $\mathbf{r}_i$ and $\Psi_e(\mathbf{r}_i)$ have different surface forms but share the same meaning.
In this paper, we consider the closed-set (e-SNLI) and multi-choice (CoS-E v1.0, CoS-E v1.11) text classification tasks.
Each task requires a different form of $\Psi_e$.

For closed-set classification, $\mathbf{r}_i$ is generally a single-word label (\eg ``entailment''), hence yielding a single-word reference rationale.
Such single-word rationales are not representative of free-text rationales, which are motivated as being accessible natural language sentences.
Here, we design $\Psi_e$ to construct $\Psi_e(\mathbf{r}_i)$ by paraphrasing $\mathbf{r}_i$ as a natural language sentence.
For each class $c_j$ in the $m$-class label space $C = \{c_j\}_{j=1}^{m}$, we manually write a set of $k$ paraphrase sentences $S_j$, based only on $c_j$ (\ie not on $\mathbf{x}_i$).
Given task instance $i$ and $\mathcal{F}$'s predicted label $y_i$, suppose $y_i = c_j$ for some $j$.
After setting $\mathbf{r}_i = y_i$, we obtain $\Psi_e(\mathbf{r}_i)$ by randomly sampling a paraphrase sentence $s$ from $S_j$ then setting $\Psi_e(\mathbf{r}_i) = s$, \eg if $\mathbf{r}_i = $ ``\texttt{contradiction}'', and $s = $ ``\texttt{The hypothesis conflicts with the premise.}'', the resulting $\Psi_e(\mathbf{r}_i)$ would be ``\texttt{The hypothesis conflicts with the premise.}''

For multi-choice classification, $\mathbf{r}_i$ is generally a multi-word choice (\eg ``have fun in the park''), hence yielding a multi-word reference rationale $\mathbf{r}_i$.
Since each instance has a different set of choices, it is not feasible to manually write paraphrase sentences specifically for every choice across all instances in the dataset.
Instead, we design $\Psi_e$ to perturb $\mathbf{r}_i$ by inserting a generic supporting sentence in front of $\mathbf{r}_i$.
First, we manually write a set, $S$, of $k$ sentences, each generically expressing affirmation for a subsequent rationale.
Note that these sentences are written independently of any particular task input, label, or rationale.
Then, for each instance $i$, we randomly sample one sentence $s$ from $S$ to insert before $\mathbf{r}_i$.
For example, if $\mathbf{r}_i = $ ``\texttt{I am a rationale.}'', and $s = $ ``\texttt{The following rationale is faithful: }'', then the resulting $\Psi_e(\mathbf{r}_i)$ would be ``\texttt{The following rationale is faithful: I am a rationale.}''

For both tasks, we can see that $\Psi_e(\mathbf{r}_i)$ is constructed in a way that does not change the meaning of $\mathbf{r}_i$.
Therefore, in both cases, a good $\Phi$ should yield a low value of $|\Phi(\mathbf{r}_i) - \Phi(\Psi_e(\mathbf{r}_i))|$.

\paragraph{Contrastive Perturbation}
Let $\Psi_c$ be an contrastive perturbation function which modifies $\mathbf{r}_i$'s surface form while also significantly changing $\mathbf{r}_i$'s semantic content.
That is, $\mathbf{r}_i$ and $\Psi_e(\mathbf{r}_i)$ should have dissimilar surface forms and meaning.

For both closed-set and multi-choice classification, we can construct $\Psi_c(\mathbf{r}_i)$ in the same way.
For each instance $i$, $y_i$ is one of $m$ classes, so we obtain $\Psi_c(\mathbf{r}_i)$ by randomly sampling one of the other $m-1$ classes to replace $\mathbf{r}_i$ with.
We can see that $\Psi_c(\mathbf{r}_i)$ is constructed in a way that fundamentally changes the meaning of $\mathbf{r}_i$.
Therefore, a good $\Phi$ should yield a high value of $|\Phi(\mathbf{r}_i) - \Phi(\Psi_c(\mathbf{r}_i))|$.

\end{document}